\documentclass[journal]{IEEEtai}

\usepackage[colorlinks,urlcolor=blue,linkcolor=blue,citecolor=blue]{hyperref}

\usepackage{color,array}

\usepackage{graphicx}

\usepackage{algorithm}
\usepackage{algorithmic}
\usepackage{multirow}
\usepackage{subfig}

\let\labelindent\relax
\usepackage{enumitem}

\usepackage{booktabs}
\usepackage{amsmath}

\begin{document}

\title{IFAR: Multi-Perspective and Multi-Level Causal Discovery with LLMs} 

\author{
Jinwei~He and Feng~Lu
\thanks{Jinwei He and Feng Lu are with State Key Laboratory of VR Technology and Systems, School of CSE, Beihang University
(e-mail: lufeng@buaa.edu.cn).}
\thanks{Corresponding author: Feng Lu.}
\thanks{This work is available as a preprint on arXiv.}
}

\maketitle

\begin{abstract}
Large language models (LLMs) have developed rapidly, and their reasoning capabilities have become a hot research topic. However, there is still limited exploration of abductive reasoning. The multi-perspective and multi-level of causes is one of the core challenges of abductive reasoning, which cannot be solved well by existing methods. We construct a specialized dataset named DeepAbduction, which is designed for tracing the causes of pollution and disease, addressing the lack of datasets in this field. We propose \textsc{Inverse-Forward Abductive Reasoning} (IFAR) framework for LLMs multi-perspective and multi-level abductive reasoning. IFAR is zero-shot and combines generalized backward reasoning with relation-by-relation forward verification. Experimental results show that IFAR achieves an improvement of approximately 40\% in the F1 score compared to other methods under mainstream LLMs, while maintaining a balance between recall and precision. Furthermore, IFAR enhances the performance of non-reasoning LLMs to surpass LLMs which have been trained for reasoning, and remains effective when applied to the latter. Code will be released after the acceptance of our work.
\end{abstract}

\begin{IEEEImpStatement}
Abductive reasoning is essential for understanding complex real-world phenomena such as pollution sources and disease transmission. However, existing language models struggle with this capability, limiting their usefulness in these analytical tasks. Our dataset and framework can improve the reliability of abductive reasoning in LLMs and significantly enhances their accuracy in tracing multi-level and multi-perspective causes. This advancement enables AI systems to better support applications in environmental monitoring, public health analysis, scientific investigation, and other domains that rely on causal understanding. By reducing expert workload and improving decision quality, our method has the potential to be applied in critical societal and industrial settings.
\end{IEEEImpStatement}


\section{Introduction}

\begin{figure}[htbp]
\centerline{\includegraphics[width=1.0\linewidth]{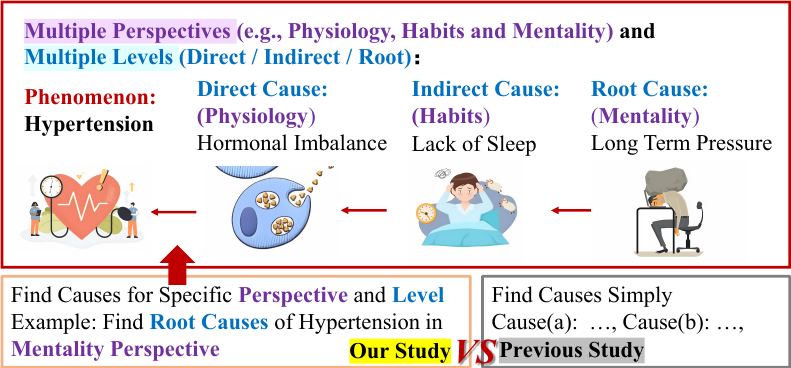}}
\caption{An example for multi-perspective and multi-level abductive reasoning. Abductive problem for previous study fails to distinguish the perspectives and reasoning levels of the causes. For eaxmple, causes of hypertension have different levels, including direct cause, indirect cause and root cause. They are also in different perspectives, such as physiology, habits and mentality perspectives. }
\label{fig44}
\end{figure}

\IEEEPARstart{T}{he} development of large language models (LLMs) marks a pivotal step toward artificial general intelligence (AGI), with research focusing on fine-tuning \cite{6(gpt),7(lora)}, agents \cite{8(trafficgpt),9(hugginggpt),tai2}, and applications \cite{1(xuanyuan),tai3,tai5}. But LLMs still fall short of true intelligence: deep reasoning ability. As a result, recent training paradigms have focused on enhancing reasoning, and developing methods to further enhance LLMs reasoning capacity has become a key research focus \cite{tai4}.

Current work is categorized into prompt‑based methods (e.g., CoT \cite{12(wei2022chain)}, ToT \cite{13(yao2024tree)}), which can be directly used yet offer limited performance gains, and reinforcement learning or fine‑tuning, which achieve the state-of-the-art reasoning performance but at the cost of substantial data and computational resources. However, all of them are mainly concerned with forward reasoning \cite{12(wei2022chain),13(yao2024tree),11(wang2022self),34(besta2024graph)}, where a more complex inverse task, called abductive reasoning, remains relatively underexplored. At present, research on abductive reasoning is still at an early stage. Most work is limited to exploring single-level abductive relations \cite{48(ma2024causal),49(joshi2024llms),tai1} or making single-cause judgments \cite{47(he2024causejudger)}. Abductive reasoning is far more complex than merely finding a cause, so what are the major challenges and their solutions?

To answer this question, we need to carefully analyze the nature of abductive reasoning, which aims to find the causes of a target phenomenon \cite{16(magnani2023handbook)}. In real situations, there are many causes that could lead to the phenomenon, all of which are plausible. These causes are often multi-level, including direct, indirect, and root causes. Their perspectives are also diverse, as an example shown in Figure \ref{fig44}. The ability to precisely identify causes as needed is one of the core challenges for abductive reasoning and it remains an area that previous research has yet to fully explore.

\begin{figure*}[htbp]
    \centering
    \includegraphics[width=1.0\linewidth]{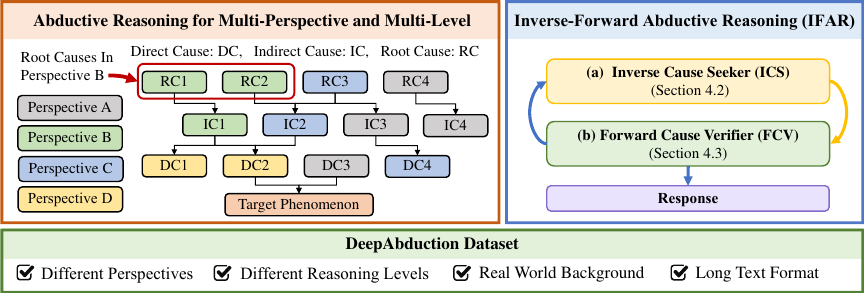}
    \caption{Overview of our study. We are the first work to analyze the challenges of multi-perspective and multi-level abductive reasoning for LLMs. To perform a detailed study of this task, we construct DeepAbduction dataset, which is specifically designed for this task. And we propose \textsc{Inverse-Forward Abductive Reasoning} (IFAR) framework for LLMs, which has two modules: (a) Inverse Cause Seeker (ICS); (b) Forward Cause Verifier (FCV).}
    \label{fig1}
\end{figure*}

Existing abductive datasets \cite{21(tafjord2021proofwriter), 22(young2022abductionrules), 47(he2024causejudger)} lack both (1) multiple reasoning levels and (2) diverse causal perspectives, making them unsuitable for studying the full complexity of abductive reasoning. To systematically examine multi-perspective and multi-level abduction, we analyzed domains where such structures naturally arise and identified two most representative real-world scenarios: \textbf{(1) pollution cause tracing} and \textbf{(2) disease cause tracing}. These domains inherently involve rich causal hierarchies and diverse viewpoints, making them ideal testbeds for multi-perspective and multi-level abductive reasoning. Based on this analysis, we construct the \textbf{DeepAbduction} dataset, specifically designed to enable research on this missing but essential aspect of abduction.

With this dataset, we can then evaluate how LLMs handle multi-perspective and multi-level abductive reasoning. The essence of a multi‑perspective and multi‑level abduction problem is to start from the observed phenomenon and, within a background that offers both high reasoning depth and diverse causal viewpoints, finding the causes that satisfy the required reasoning level and perspective. We claim that this task is intrinsically challenging and even LLMs explicitly trained for deep reasoning still struggle to solve it perfectly. We also argue that, with a well‑devised strategy, even non-reasoning type LLMs can achieve strong performance. 

To address this, we draw inspiration from human reasoning processes\cite{45(tt),46(cc)}: (1) humans often use backward and forward thinking simultaneously when reasoning; (2) humans are skilled in solving complex reasoning step by step. We propose the \textbf{\textsc{Inverse-Forward Abductive Reasoning} (IFAR)} abductive reasoning framework, which combines divergent backward reasoning with precise forward verification. IFAR consists of two modules: (1) \textbf{Inverse Cause Seeker (ICS)} and \textbf{Forward Cause Verifier (FCV)}. The overall structure of our approach is illustrated in Figure \ref{fig1}. 

In the experimental section, on non-reasoning type LLMs, our method outperforms the comparison methods, and even outperforms the LLMs explicitly trained for reasoning, achieving a balance between recall and precision. 

In summary, our contributions to the field are threefold:

\begin{itemize}[leftmargin=*]
    \item \textbf{Novel Task}: To the best of our knowledge, we are the first to define, analyze and address the \textit{multi-perspective and multi-level challenges} in LLMs abductive reasoning. 

    \item \textbf{Foundational Dataset}: We construct \textit{DeepAbduction}, the first dataset that simultaneously provides multiple causal perspectives and multiple reasoning levels, with real-world background, addressing the lack of dataset in this field. 

    \item \textbf{Effective Solution}: We propose \textit{IFAR framework}, which is zero-shot and can balance the recall and precision, outperforming comparison methods and LLMs explicitly trained for reasoning in the experiments.
\end{itemize}

\begin{figure*}[htbp]
\centering
  \includegraphics[width=1.0\linewidth]{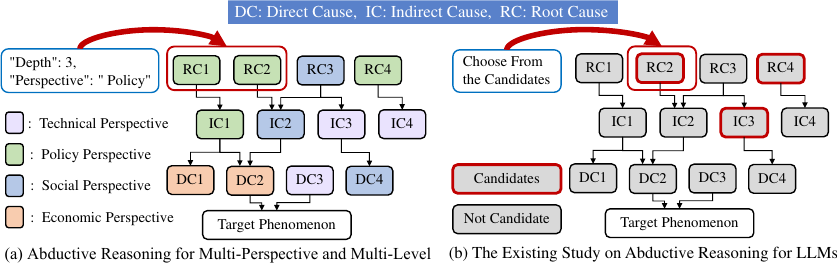}
	\caption{The differences between existing research and the multi-perspective and multi-level abductive reasoning proposed in this paper. The different colors represent different perspectives, where (a) represents multi-perspective and multi-level abductive reasoning, and (b) represents the existing task which is limited to single perspective and based on candidate causes.}
	\label{fig2}
\end{figure*}

\section{Related Works}
\paragraph{Large Language Models Reasoning}
The appearance of LLMs \cite{23(touvron2023llama),26(zeng2022glm)} has marked a new wave of excitement in artificial intelligence. LLMs possess strong language understanding and reasoning capabilities \cite{10(large)}, but studies have shown that their performance on specific logical problems is still underdeveloped \cite{29(xu2023large),30(huang2023towards)}. The current training methods for LLMs are developing in the direction of enhancing reasoning. The latest reasoning LLMs include Deepseek-R1 \cite{53(guo2025deepseek)}, OpenAI's o-series models \cite{43(openai-o3)}, etc. Among them, Deepseek-R1 uses reinforcement learning, allowing LLMs to learn the thinking patterns of Chain-of-Thought \cite{12(wei2022chain)} method, enabling deep thinking capabilities. Pre-training or fine-tuning requires data and computational resources, and is not applicable to non-open source models. Therefore, strengthening the reasoning capabilities of LLMs without altering parameters has become a key research area \cite{35(yu2023thought),36(lin2024swiftsage)}. Through the study and simulation of human thinking processes, researchers have designed reasoning frameworks for LLMs to enhance their ability to complete complex tasks \cite{13(yao2024tree),35(yu2023thought)}. Chain-of-thought (CoT) \cite{12(wei2022chain)} stimulates the chain reasoning of LLMs by adding "Let's think step by step" to the input prompt. Self-Consistency Chain-of-thought (SC-CoT) \cite{11(wang2022self)} applies the idea of multiple rounds of thinking, answers are collected from various reasoning paths, and the most consistent one is ultimately selected as the final answer. Information Re-Organization (InfoRE) \cite{50(cheng2024information)} reorganizes the text before reasoning to solve reasoning problems with long context.

\paragraph{Abductive Reasoning}
Abductive reasoning is a reverse reasoning process that aims to find an explanation that satisfies the requirements for a target result or phenomenon \cite{37(thagard1997abductive),38(josephson1996abductive),17(bhagavatula2020abductive)}. Abductive reasoning involves reasoning based on causal chains and mainly relies on the logical ability of the model. Classic abductive reasoning datasets include ProofWriter \cite{21(tafjord2021proofwriter)}, AbductiveRules \cite{22(young2022abductionrules)} and CauseLogics \cite{47(he2024causejudger)}, which all have complex logical structures and are highly challenging, but the data items only consider a single perspective and lack consideration of multiple perspectives. In the field of language models, research on abductive reasoning mainly focuses on small model training or fine-tuning \cite{39(bai2023abductive),40(aakur2019abductive),41(huang2020semi)}. As for abductive reasoning methods for LLMs, they are currently limited to judging the correctness of a given cause \cite{47(he2024causejudger)}, which fails to reflect the core of abductive reasoning. Searching for causes at specific level and perspective is still an unsolved research problem.

\section{Motivation and Dataset}
\subsection{Abduction with Multi-Perspective and Multi-Level}
Currently, in the classical research on reasoning in LLMs, most studies focus on deductive and forward reasoning problems \cite{10(large),11(wang2022self),12(wei2022chain)}, with comparatively less attention given to abductive reasoning. Among the few studies addressing abductive reasoning \cite{47(he2024causejudger),21(tafjord2021proofwriter),22(young2022abductionrules)}, there are two common flaws: (1) They do not differentiate between perspectives and often reduce the cause to a single perspective, thereby failing to capture the multi-perspective nature of the abductive process; (2) They are based on candidate set with only one correct cause and treats it as the definitive truth, \textbf{neglecting the multi-possibility nature} of the abductive process. 

Part (b) of Figure \ref{fig2} represents the existing tasks, where the valid reasoning chains are limited and it can only reflect a single perspective. In real-world, causes are multi-level, and multiple perspectives are involved. Here, level represents the shortest logical hops between two nodes. Depending on the situation, the causes people focus on may vary in terms of perspectives and levels. Technical personnel tend to focus on technical causes, government officials often focus on policy-related causes, as well as implementers concentrate on direct causes, while leaders generally focus on higher-level causes. Abductive reasoning for specific causes from a specific perspective and level presents a significant challenge in abductive reasoning, as shown in Figure \ref{fig2} (a). Due to the high difficulty and complexity of these issues, relying solely on the capabilities of LLMs is insufficient. Therefore, there is an urgent need to explore multi-perspective and multi-level abductive reasoning methods for LLMs.

\begin{figure*}[htbp]
\centering
  \includegraphics[width=1.0\linewidth]{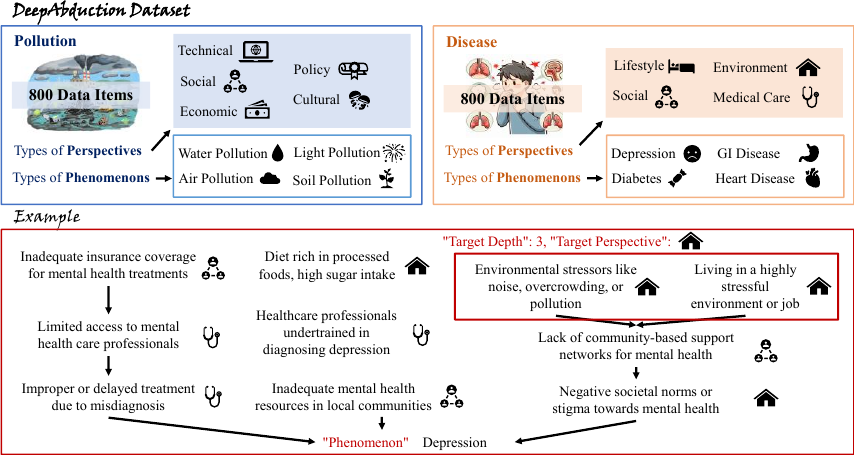}
	\caption{The overview of our proposed DeepAbduction dataset. DeepAbduction includes two representative abductive reasoning scenarios: \textbf{pollution} and \textbf{disease}, which encompass a rich variety of phenomenons and perspectives. The content in our dataset contains multi-level and multi-perspective causal relationships, requiring simultaneous consideration of both reasoning levels and perspectives to obtain the target causes.}
	\label{figCaseDataset}
\end{figure*}

\begin{table*}[htbp]
\caption{A sample of DeepAbduction, which is designed for multi-perspective and multi-level abductive reasoning tasks.}
\renewcommand{\arraystretch}{1}
\centering
\begin{tabular}{l}
\toprule
\textbf{"Premise Text"}: "Low self-esteem and poor mental health management skills are often associated with living in areas with high crime rates or frequent \\ violence. Environmental stressors such as noise, overcrowding, or pollution can lead to limited access to recreational spaces or green areas. High costs \\ and travel difficulties act as barriers to accessing mental health care, contributing to excessive work hours and eventually  leading to burnout. Living in \\ high-crime or violent areas is also linked to poor housing conditions or a lack of stable housing, as well as overcrowded, noisy, or polluted environments. \\

A lack of time for self-care or relaxation reinforces negative societal norms or stigma surrounding mental health. Insufficient training among healthcare \\ professionals in diagnosing depression also contributes to stigma around seeking mental health support. In addition, lack of physical activity and exercise, \\ negative societal norms regarding mental health, insufficient workplace mental health support or benefits, discrimination and inequality in accessing \\ mental health services, and poor housing conditions or unstable living situations can all lead to depression. ……"\\
\textbf{"Perspectives"}: \\
"Personal Lifestyle": ["Lack of time for self-care or relaxation", "Low self-esteem and poor mental health management skills", "Excessive work hours \\ leading to burnout", ……], \\
"Living Environment": ["Environmental stressors like noise, overcrowding, or pollution", "Living in areas with high crime rates or violence", "Poor \\ housing conditions or lack of a stable home", ……], \\
"Medical Care": ["Barriers to accessing mental health care, such as high costs or travel difficulties", ……] , \\
"Socioeconomic": ["Lack of workplace mental health support or benefits", …… ]\\
\textbf{"Phenomenon"}: "Depression", \\ 
\textbf{"Target Depth"}: 3, \\
\textbf{"Target Perspective"}: "Personal Lifestyle", \\
\textbf{"Label"}: "Low self-esteem and poor mental health management skills", …… \\
\bottomrule
\end{tabular}
\label{table1}
\end{table*}

\subsection{DeepAbduction Dataset}
To investigate this problem, a suitable dataset is indispensable. However, existing resources do not support multi-perspective and multi-level. Therefore, we construct the DeepAbduction dataset, designed around two real-world backgrounds that naturally exhibit rich causal hierarchies: \textbf{(1) pollution} and \textbf{(2) disease}. Each background contains eight sub-themes, and each sub-theme includes 200 instances, resulting in a total of 1,600 samples. Figure \ref{figCaseDataset} presents a overview of our dataset. The content in  of this dataset contains multi-level and multi-perspective causal relationships.

The causes involved are based on real-world network materials, summarized through LLMs and manually verified to ensure they reflect real-world attributes. A sample of the dataset structure is shown in Table \ref{table1}. "Phenomenon" refers to the target phenomenon to be explained (e.g., Depression).
"Target Depth" represents the needed level of abductive reasoning for the target phenomenon, where the causes directly inferred in one step are level 1, and so on.
"Target Perspective" represents to the needed perspective of the cause, such as living environment, medical care condition, etc. The task is to infer the causes of the target phenomenon that meet the given target perspective and depth.
"Premise Text" represents a text which contains all abductive relationships.
"Perspectives" provides the perspective labels for each cause in the text.
"Label" is the correct answer, which may have more than one cause, usually between 1 and 3 causes. 

\begin{figure*}[t]
	\centering
	\includegraphics[width=1.00\linewidth]{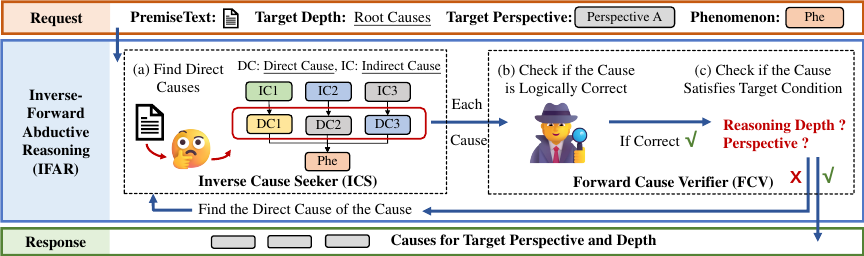}
	\caption{Overview of our proposed \textsc{Inverse-Forward Abductive Reasoning} (IFAR) framework. Different colors represent different perspectives. IFAR is divided into two modules, Inverse Cause Seeker (ICS) and Forward Cause Verifier (FCV).}
	\label{fig3}
\end{figure*}

The statistics of DeepAbduction are shown in Table \ref{tabledata}. Its advantages are as follows. (1) Our dataset is specifically designed for the task of multi-perspective and multi-level abductive reasoning. To the best of our knowledge, we are the first to introduce the concept of multi-perspective in the field of abductive reasoning datasets. (2) Our dataset includes a wide range of causes and perspectives, enhancing its comprehensiveness and complexity. Moreover, it is presented in textual form, which more closely reflects real-world scenarios compared to directly providing relational pairs. (3) Based on the real-world materials, our dataset possesses practical context with two classical abductive scenarios, offering an advantage over datasets that are entirely artificially generated. 

Finally, it is important to note that while our dataset is built upon real-world contexts, it is specifically designed for AI abductive reasoning. As such, the construction prioritizes well-defined logical structures over comprehensive empirical rigor. Consequently, while the scenarios provide meaningful contexts for reasoning, the data still lacks the expert validation required for professional research in fields such as environmental pollution or clinical medicine. Users are advised to keep this distinction clearly in mind when applying the dataset.

\begin{table}[htbp]
 \caption{Statistics of DeepAbduction Dataset.}
  \centering
     \begin{tabular}{p{4cm}|p{1.3cm}p{1.3cm}}
     \toprule
    \textbf{Dataset Themes} &
    \textbf{Pollution} & \textbf{Disease} \\ 
    \midrule
    Amount of Data Items & 800 & 800 \\
    Types of Perspectives & 5 & 4  \\
    Types of Phenomenons & 4 & 4 \\
    Max Relation Depth & 4 & 4 \\
    Avg Text Length & 1452 & 1306 \\
    Avg Num of Relations & 13 & 11 \\
    Avg Num of Causes & 22 & 19 \\
    \bottomrule
  \end{tabular}
  \label{tabledata}
\end{table}

\begin{table*}[htbp]
  \caption{Experimental results on mainstream LLMs which are not trained for reasoning. Our proposed IFAR outperforms existing methods across all three LLMs, while maintaining a balance between recall and precision score. By simply using ICS and adding FCV to IO, we discuss the performance of ICS and FCV individually, as well as their combined interaction.}
  \centering
    \begin{tabular}{c|c|l|ccc}
    \toprule
    \textbf{Dataset} & \textbf{LLMs-Backend} & \textbf{Method} &
    \textbf{F1 Score ($\uparrow$)} & \textbf{Recall (\%) ($\uparrow$)} & \textbf{Precision (\%) ($\uparrow$)} \\ 
    \midrule
    \multirow{19.5}{*}{\textbf{DeepAbduction-Pollution}} &
    \multirow{7}{*}{\textbf{GLM-4-Air} \cite{44(glm-4-air)}} &
    IO \cite{10(large)} & 0.30 & 43.3 & 22.7  \\
    & & Chain-of-thought (CoT) \cite{12(wei2022chain)} & 0.32 & 49.0 & 23.6  \\
    & & Self-Consistency CoT (SC-CoT) \cite{11(wang2022self)} & 0.34 & 44.7 & 27.6 \\
    \addlinespace[1pt]
    \cline{3-6}
    \addlinespace[1pt]
    & & IO and FCV (Ours) & 0.33 & 41.3 & 27.1 \\
    & & Only ICS (Ours) & 0.65 & 92.0 & 50.5 \\
    & & \textbf{IFAR (Ours)} & \textbf{0.76} & \textbf{73.9} & \textbf{78.9} \\
    \addlinespace[1pt]
    \cline{2-6}
    \addlinespace[1pt]
    & \multirow{7}{*}{\textbf{Qwen-2.5-14B} \cite{52(yang2024qwen2)}}
    & IO \cite{10(large)} & 0.39 & 48.7 & 31.9 \\
    & & Chain-of-thought (CoT) \cite{12(wei2022chain)} & 0.40 & 53.3 & 31.9 \\
    & & Self-Consistency CoT (SC-CoT) \cite{11(wang2022self)} & 0.44 & 60.7 & 34.0 \\
    \addlinespace[1pt]
    \cline{3-6}
    \addlinespace[1pt]
    & & IO and FCV (Ours) & 0.50 & 44.7 & 56.8 \\
    & & Only ICS (Ours) & 0.67 & 82.7 & 56.6 \\
    & & \textbf{IFAR (Ours)} & \textbf{0.79} & \textbf{70.7} & \textbf{89.8} \\
    \addlinespace[1pt]
    \cline{2-6}
    \addlinespace[1pt]
    & \multirow{7}{*}{\textbf{Deepseek-V3} \cite{51(liu2024deepseek)}} &
    IO \cite{10(large)} & 0.51 & 75.3 & 38.8 \\
    & & Chain-of-thought (CoT) \cite{12(wei2022chain)} & 0.52 & 85.3 & 37.5 \\
    & & Self-Consistency CoT (SC-CoT) \cite{11(wang2022self)} & 0.52 & 85.3 & 37.5 \\
    \addlinespace[1pt]
    \cline{3-6}
    \addlinespace[1pt]
    & & IO and FCV (Ours) & 0.53 & 67.3 & 43.9 \\
    & & Only ICS (Ours) & 0.91 & 94.7 & 88.2 \\
    & & \textbf{IFAR (Ours)} & \textbf{0.93} & \textbf{87.3} & \textbf{98.5} \\
    \midrule
    \multirow{19.5}{*}{\textbf{DeepAbduction-Disease}} &
    \multirow{7}{*}{\textbf{GLM-4-Air} \cite{44(glm-4-air)}} &
    IO \cite{10(large)} & 0.44 & 60.0 & 34.6  \\
    & & Chain-of-thought (CoT) \cite{12(wei2022chain)} & 0.44 & 62.6 & 34.2  \\
    & & Self-Consistency CoT (SC-CoT) \cite{11(wang2022self)} & 0.47 & 71.6 & 34.8 \\
    \addlinespace[1pt]
    \cline{3-6}
    \addlinespace[1pt]
    & & IO and FCV (Ours) & 0.51 & 52.9 & 49.1 \\
    & & Only ICS (Ours) & 0.70 & 88.4 & 58.1 \\
    & & \textbf{IFAR (Ours)} & \textbf{0.75} & \textbf{62.6} & \textbf{95.1} \\
    \addlinespace[1pt]
    \cline{2-6}
    \addlinespace[1pt]
    & \multirow{7}{*}{\textbf{Qwen-2.5-14B} \cite{52(yang2024qwen2)}}
    & IO \cite{10(large)} & 0.50 & 69.7 & 39.0 \\
    & & Chain-of-thought (CoT) \cite{12(wei2022chain)} & 0.51 & 74.2 & 39.0 \\
    & & Self-Consistency CoT (SC-CoT) \cite{11(wang2022self)} & 0.52 & 80.6 & 38.1 \\
    \addlinespace[1pt]
    \cline{3-6}
    \addlinespace[1pt]
    & & IO and FCV (Ours) & 0.53 & 60.0 & 47.7 \\
    & & Only ICS (Ours) & 0.71 & 67.7 & 74.5 \\
    & & \textbf{IFAR (Ours)} & \textbf{0.77} & \textbf{62.6} & \textbf{99.0} \\
    \addlinespace[1pt]
    \cline{2-6}
    \addlinespace[1pt]
    & \multirow{7}{*}{\textbf{Deepseek-V3} \cite{51(liu2024deepseek)}} &
    IO \cite{10(large)} & 0.53 & 94.2 & 36.8 \\
    & & Chain-of-thought (CoT) \cite{12(wei2022chain)} & 0.52 & 92.3 & 36.7 \\
    & & Self-Consistency CoT (SC-CoT) \cite{11(wang2022self)} & 0.51 & 92.9 & 35.5 \\
    \addlinespace[1pt]
    \cline{3-6}
    \addlinespace[1pt]
    & & IO and FCV (Ours) & 0.50 & 54.2 & 46.9 \\
    & & Only ICS (Ours) & 0.91 & 93.5 & 87.9 \\
    & & \textbf{IFAR (Ours)} & \textbf{0.94} & \textbf{88.7} & \textbf{99.3} \\
    \bottomrule
  \end{tabular}
  \label{table2}
\end{table*}

\subsection{Dataset Construction}
\label{appendix3}
The construction process of DeepAbduction is organized into three stages: causal relation extraction, relation integration with structured labeling, and controlled instance construction.

\subsubsection{Causal Relation Extraction from Online Materials}

We first collect domain-relevant online materials through curated keyword searches. Each document is segmented into fine-grained textual units (e.g., paragraphs or bullet points). For each unit, a LLM is instructed to extract explicit causal relations in the form of ⟨cause, effect⟩ pairs. This ensures that all extracted relations are grounded in real text rather than generated freely. All extracted relations are stored with their source identifiers to maintain traceability.

\subsubsection{Relation Integration, Layer Structuring, and Perspective Labeling}

The extracted causal pairs from all sources are then merged and organized into a unified causal graph centered on the topic. This stage includes three structured steps: Layer assignment: Relations are grouped into multiple reasoning layers, including direct causes, intermediate causes (second- and third-level), and root causes. The layer assignment is determined by comparing abstraction levels and causal dependency chains between relations. Perspective labeling: Each relation is annotated with one or more abductive perspectives (e.g., social, psychological, economic, technical), which are determined based on content semantics and predefined perspective criteria.

\subsubsection{Quality control}
Quality control is conducted through a two-stage filtering mechanism designed to ensure the reliability, internal consistency, and factual grounding of the constructed causal graph.

\textbf{LLM Consistency Verification.} To mitigate extraction noise and reduce hallucination, each textual unit is processed by the extraction model $T$ times (typically $T=3$--$5$). Only relations consistently extracted across multiple passes are retained. This step filters one-off hallucinations, low-confidence relations, and inconsistent causal directions. Redundant or semantically equivalent relations are further merged using embedding-based similarity checking.

\subsubsection{Dataset Construction from the Causal Graph}

To construct each data sample, we select a subset of causes from the integrated causal graph according to predefined settings, such as target reasoning depth or intended perspective.

For each selected partial cause set, the LLM is prompted to generate an abductive explanation that connects the chosen subset to the overall causal chain.
Each resulting instance contains: the selected causes as the reasoning anchor, the corresponding target reasoning depth or perspective, and the generated explanation grounded in the structured causal relations. This guarantees that each data item is constructed directly from the underlying causal graph and reflects controllable and interpretable abductive reasoning patterns.

\begin{algorithm}[ht]
    \caption{IFAR(\(T_{\text{Premise}}\),\ \(Phe\),\ \(D_{\text{Target}}\),\ \(P_{\text{Target}}\))}
    \label{al1}
    {\bf Input:} \\ Text \(T_{\text{Premise}}\), Target phenomenon \(Phe\), Target depth \(D_{\text{Target}}\), Target perspective \(P_{\text{Target}}\)\\
    {\bf Output:}   
    \begin{algorithmic}[1]
    \STATE \(List_{\text{Phe}} \gets [ Phe ] \)
    \STATE \(\mathcal{C_{\text{Valid}}} \gets \phi\)
    \WHILE{\(List_{\text{Phe}} \mathrm{\ is \ not \ empty} \)}
    \STATE \(phe \gets List_{\text{Phe}}.get() \)
    \STATE \(\mathcal{C_{\text{ICS}}} = \mathrm{ICS^{LLM}}(T_{\text{Premise}}, phe, p_{\text{ICS}})\)
    \FOR{\(C \in \mathcal{C_{\text{ICS}}}\)}
    \STATE{\(V_{\text{IsDirectCause}} = \mathrm{FCV^{LLM}_{S1}}(C, phe, D_{\text{T}}, P_{\text{T}}, p_{\text{1}})\)}
    \STATE{\(V_{\text{IsTarget}}, D_{\text{Current}} = \mathrm{FCV^{LLM}_{S2}}(C, phe, D_{\text{T}}, P_{\text{T}}, p_{\text{2}})\)}
    \IF{\(V_{\text{IsDirectCause}}\) and \(V_{\text{IsTarget}}\)}
    \STATE \(\mathcal{C_{\text{Valid}}}.add(C)\)
    \ELSIF{\(V_{\text{IsDirectCause}}\) and \(D_{\text{Current}} < D_{\text{Target}}\)}
    \STATE \(List_{\text{Phe}}.put(C)\)
    \ENDIF
    \ENDFOR
    \ENDWHILE
    \RETURN \(\mathcal{C_{\text{Valid}}}\)
    \end{algorithmic}
\end{algorithm}

\section{IFAR Framework}
\subsection{Overall Design}
 We propose a multi-perspective and multi-level abductive reasoning framework for LLMs, named \textbf{\textsc{Inverse-Forward Abductive Reasoning} (IFAR)}, which comprises two core modules: Inverse Cause Seeker (ICS) and Forward Cause Verifier (FCV), as shown in Figure \ref{fig3}. The specific process of IFAR is as follows: First, the ICS module identifies all direct causes of the current phenomenon in a single query, utilizing a divergent search strategy. Then, for each cause, the FCV module conducts a forward validation to assess its accuracy and determine whether the cause satisfies the required level and perspective. If the cause meets the criteria, it is added to the answer set. Otherwise, it is treated as a new phenomenon and the ICS module is used again to identify the direct causes of it, continuing the process of deepening reasoning.

\subsection{Inverse Cause Seeker}
Abductive reasoning is a reverse process, where multiple causes may lead to the target phenomenon. It is difficult for LLMs to obtain the correct answer directly while considering both levels and perspectives simultaneously. It is easier for LLMs to firstly identify the direct cause of the current phenomenon. If we only look for one cause in a single step, it will greatly reduce efficiency. Therefore, the ICS aims to find all direct causes of the phenomenon in a single step:

\begin{equation}
\mathcal{C_{\text{ICS}}} = \mathrm{ICS^{LLM}}(T_{\text{Premise}}, phe, p_{\text{ICS}})
\end{equation}

where \( T_{\text{Premise}} \) represents premise text, \( phe \) is the phenomenon, \(D_{\text{ICS}}\) denote the reasoning depth is 1, \(p_{\text{ICS}}\) is structured prompt. \(\mathrm{ICS^{LLM}}\) is ICS method which calls LLMs to find the direct causes of the current phenomenon in one step. \(\mathcal{C_{\text{ICS}}}\) is the set of direct causes discovered by ICS.

\begin{figure*}[t]
	\centering
	\includegraphics[width=1.0\linewidth]{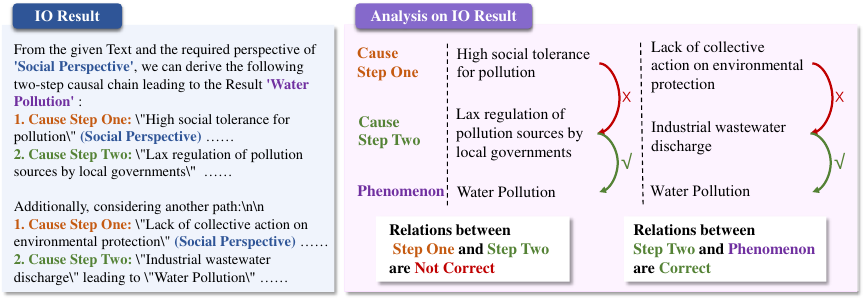}
	\caption{An example that errors occur in IO method. The single-level reasoning (green part) is logically correct in IO method, while the multi-level reasoning (orange part) has hallucinations.}
	\label{fig6}
\end{figure*}

\begin{table*}[htbp]
  \caption{ The advantage of IFAR over classical methods surpasses the improvement of DsR1 over DsV3 in reasoning.}
  \centering
     \begin{tabular}{l|l|ccc}
     \toprule
    \textbf{Dataset} & \textbf{Method} &
    \textbf{F1 Score ($\uparrow$)} & \textbf{Recall (\%) ($\uparrow$)} & \textbf{Precision (\%) ($\uparrow$)} \\ 
    \midrule
    \multirow{9}{*}{\textbf{DeepAbduction-Pollution}} &
    Deepseek-V3 (DsV3) \cite{51(liu2024deepseek)} & 0.51 & 75.3 & 38.8 \\
    & Deepseek-R1 (DsR1) \cite{53(guo2025deepseek)} & 0.84 & 99.3 & 72.3 \\
    \addlinespace[1pt]
    \cline{2-5}
    \addlinespace[1pt]
    & DsR1 \cite{53(guo2025deepseek)} + Chain-of-thought (CoT) \cite{12(wei2022chain)} & 0.84 & 98.0 & 72.8  \\
    & DsR1 \cite{53(guo2025deepseek)} + Self-Consistency CoT (SC-CoT) \cite{11(wang2022self)} & 0.92 & 100.0 & 85.1 \\
    \addlinespace[1pt]
    \cline{2-5}
    \addlinespace[1pt]
    & \textit{\textbf{DsV3 \cite{51(liu2024deepseek)} + IFAR (Ours)}} & \textit{\textbf{0.93}} & \textit{\textbf{87.3}} & \textit{\textbf{98.5}} \\
    \addlinespace[1pt]
    \cline{2-5}
    \addlinespace[1pt]
    & DsR1 \cite{53(guo2025deepseek)} + FCV (Ours) & 0.97 & 94.0 & 99.3 \\
    & DsR1 \cite{53(guo2025deepseek)} + ICS (Ours) & 0.99 & 97.7 & 99.3 \\
    & \textbf{DsR1 \cite{53(guo2025deepseek)} + IFAR (Ours)} & \textbf{1.00} & \textbf{100.0} & \textbf{100.0} \\
    \midrule
    \multirow{9}{*}{\textbf{DeepAbduction-Disease}} & Deepseek-V3 (DsV3) \cite{51(liu2024deepseek)} & 0.53 & 94.2 & 36.8 \\
    & Deepseek-R1 (DsR1) \cite{53(guo2025deepseek)} & 0.84 & 98.7 & 73.7  \\
    \addlinespace[1pt]
    \cline{2-5}
    \addlinespace[1pt]
    & DsR1 \cite{53(guo2025deepseek)} + Chain-of-thought (CoT) \cite{12(wei2022chain)} & 0.87 & 96.8 & 78.5  \\
    & DsR1 \cite{53(guo2025deepseek)} + Self-Consistency CoT (SC-CoT) \cite{11(wang2022self)} & 0.89 & 92.2 & 86.7 \\
    \addlinespace[1pt]
    \cline{2-5}
    \addlinespace[1pt]
    & \textit{\textbf{DsV3 \cite{51(liu2024deepseek)} + IFAR (Ours)}} & \textit{\textbf{0.94}} & \textit{\textbf{88.7}} & \textit{\textbf{99.3}} \\
    \addlinespace[1pt]
    \cline{2-5}
    \addlinespace[1pt]
    & DsR1 \cite{53(guo2025deepseek)} + FCV (Ours) & 0.97 & 95.5 & 99.3 \\
    & DsR1 \cite{53(guo2025deepseek)} + ICS (Ours) & 0.97 & 95.5 & 98.7 \\
    & \textbf{DsR1 \cite{53(guo2025deepseek)} + IFAR (Ours)} & \textbf{0.99} & \textbf{98.7} & \textbf{100.0} \\
    \bottomrule
  \end{tabular}
  \label{tabler1}
\end{table*}

\subsection{Forward Cause Verifier}
ICS can effectively identify direct causes. However, some of the identified causes may be incorrect and require further verification. FCV performs a forward abductive relationship verification for each cause to confirm whether it is a direct cause to the current phenomenon:

\begin{equation}
\mathcal{C_{\text{Direct}}} = \{ C \in \mathcal{C_{\text{ICS}}} \mid \mathrm{FCV^{LLM}_{S1}}(C, phe, D_{\text{T}}, P_{\text{T}}, p_{\text{1}})\}
\end{equation}

And whether it satisfies the target depth and target perspective constraints:

\begin{equation}
\mathcal{C_{\text{Valid}}} = \{ C \in \mathcal{C_{\text{Direct}}} \mid \mathrm{FCV^{LLM}_{S2}}(C, phe, D_{\text{T}}, P_{\text{T}}, p_{\text{2}})\}
\end{equation}

where \(D_{\text{T}}\) is the target reasoning depth, \(P_{\text{T}}\) is the target perspective, \(p_{\text{1}}\) and \(p_{\text{2}}\) is the structured prompt. \(\mathrm{FCV^{LLM}}\) is FCV method which calls LLMs to do two-step verification of the causes from ICS. \(\mathcal{C_{\text{Direct}}}\) is the set of direct causes verified by FCV and \(\mathcal{C_{\text{Valid}}}\) is the set of causes that satisfied the target depth and perspective judged by FCV.

If the cause satisfies the criteria, it is a valid result. If not, the causes in \(\mathcal{C_{\text{Direct}}} \setminus \mathcal{C_{\text{Valid}}}\) will be treated as the next \(phe\), and a deeper search for direct causes will be performed. The process of IFAR is illustrated in Algorithm \ref{al1}. It’s worth noting that our approach only provides the LLMs with the premise text and the targets, as input. The model must itself uncover the detailed reasoning chain in the text and the perspective of each cause, which enhances the practicality.

\section{Experiments}
\subsection{Settings and Baselines}
LLMs used in the experiments are GLM-4-Air \cite{44(glm-4-air)}, Qwen-2.5-14B \cite{52(yang2024qwen2)}, Deepseek-V3 \cite{51(liu2024deepseek)} and Deepseek-R1 \cite{53(guo2025deepseek)}. GLM-4-Air and Deepseek-V3 represent non-reasoning type LLMs with a large scale of parameters. Qwen-2.5-14B, which has a slightly smaller parameter size, is one of the most widely used open-source LLMs. Deepseek-R1, on the other hand, is an LLM trained for deep reasoning through reinforcement learning. We use our proposed DeepAbduction dataset, and select 160 data points from both themes. 

\begin{figure}[htbp]
\centering
    \includegraphics[width=1.0\linewidth]{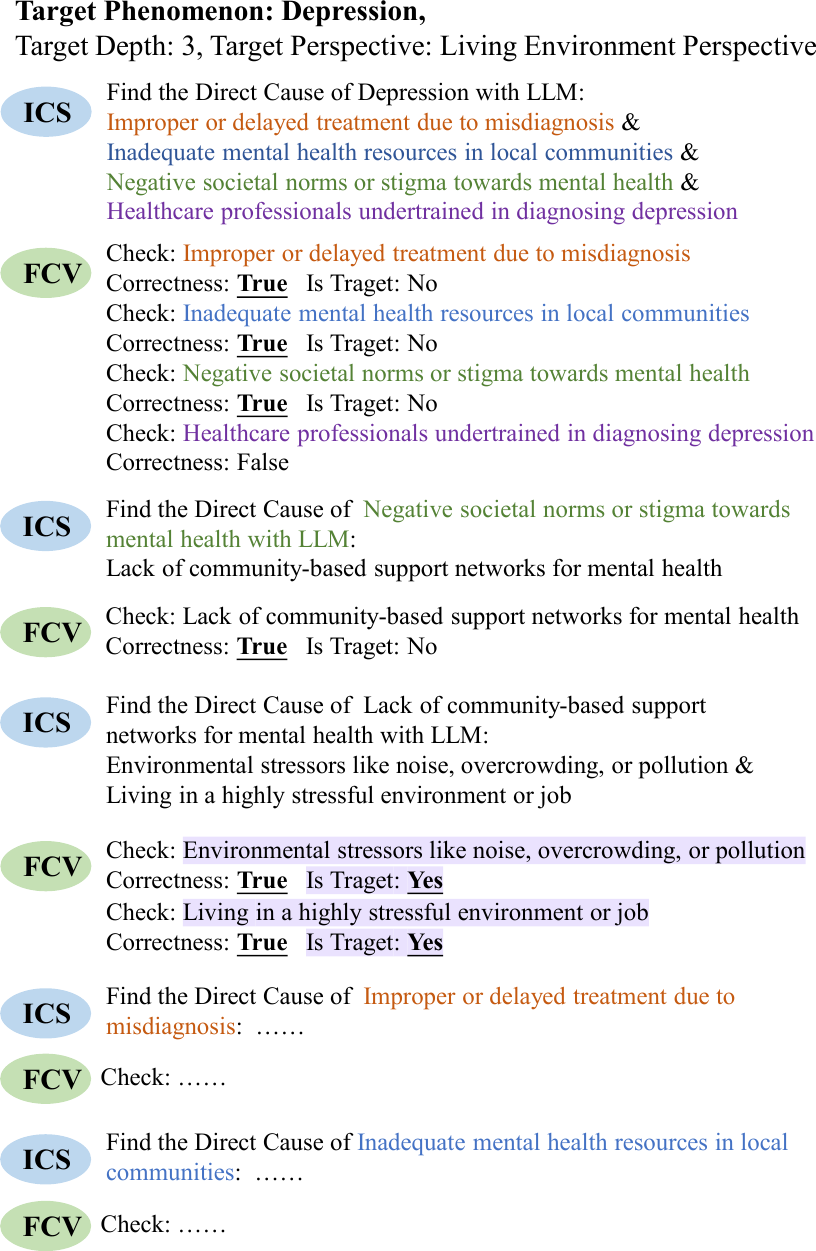}
    \caption{A case of our IFAR framework. The ICS module determines the direct cause, while the FCV module verifies the correctness of the cause from ICS and determines whether it aligns with the target depth and perspective. The ICS module and FCV module alternately operate, iteratively reasoning and verifying towards the final answer.}
    \label{fig10}
\end{figure}

The experimental metrics are recall and precision. Recall represents the proportion of correctly identified positive instances out of all actual positive instances. Precision represents the proportion of correctly identified positive instances out of all predicted positive instances. The F1 score, which is the harmonic mean of the recall and precision, is used to evaluate the overall performance. The comparative methods include IO \cite{10(large)}, Chain-of-thought (CoT) \cite{12(wei2022chain)}, Self-Consistency CoT (SC-CoT) \cite{11(wang2022self)}. The IO method involves directly using LLMs for question-and-answer. CoT and SC-CoT are classic LLM reasoning methods. CoT stimulates a chain-of-thought reasoning process, whereas SC-CoT involves multiple executions of CoT followed by a voting integration.

For GLM-4-Air \cite{44(glm-4-air)}, Deepseek-V3 \cite{51(liu2024deepseek)}, and Deepseek-R1 \cite{53(guo2025deepseek)}, we run them using the official APIs from GLM and Deepseek company, and remote API usage does not require additional compute resources. For Qwen-2.5-14B \cite{52(yang2024qwen2)}, we run it locally on the computer, with the operating system being Ubuntu 18.04.6, and the hardware resources consisting of two NVIDIA GeForce RTX 3090s. 

\subsection{Results}
\paragraph{Main Results}
Experimental results on foundational LLMs which are not explicitly trained for deep reasoning are presented in Table \ref{table2}. As shown, our method effectively improves F1 scores, while also performing well in both precision and recall, achieving a good balance (e.g., 0.93 for F1 score, 87.3 for recall, 98.5 for precision on DeepAbduction-Pollution dataset and 0.94 for F1 score, 88.7 for recall, 99.3 for precision on DeepAbduction-Disease dataset with Deepseek-V3). Notably, in terms of precision, our method significantly outperforms all other methods across all three LLMs, underscoring the effectiveness of the FCV module. IO, CoT, and SC-CoT perform not well, which indicates that the classical LLM reasoning methods have limitations in performance for multi-perspective and multi-level abductive tasks.

\paragraph{Comparison with Reasoning LLMs} As above, Deepseek-V3 (DsV3) performs the best. We believe that Deepseek-R1 (DsR1) is also a strong representative of reasoning LLMs, so we chose it for our following experiments. Results are shown in Table \ref{tabler1}. DsR1 itself achieves good results, greatly outperforming DsV3. \textbf{However, when \textit{DsV3 applied IFAR}, its performance has already surpassed DsR1 and even outperformed DsR1 combined with existing methods (e.g., IO, CoT, SC-CoT)}. This demonstrates that the performance gap between existing methods and our IFAR exceeds the reasoning efficiency advantage of DsR1 over DsV3. DsR1 has been trained through reinforcement learning specifically for reasoning, whereas our method does not require training process but still achieves better results on non-reasoning type LLMs. Furthermore, we also test DsR1 combined with our method, which also shows an improvement, reaching near-perfect performance. This demonstrates that: (1) for typical multi-perspective and multi-level abductive reasoning tasks, does not require the most advanced reasoning LLMs; \textbf{a standard based LLM combined with our IFAR can achieve good results}; (2) for cases where accuracy is critical, \textbf{using an reasoning LLM combined with IFAR can further enhance the performance}.

\begin{table*}[htbp]
  \centering
      \caption{ The performance of IFAR and CoT method across different target levels, with Deepseek-V3.}
     \begin{tabular}{c|c|c|ccc}
     \toprule
     \textbf{Method} & \textbf{Dataset} & 
    \textbf{Target Level} &
    \textbf{F1 Score ($\uparrow$)} & \textbf{Recall (\%) ($\uparrow$)} & \textbf{Precision (\%) ($\uparrow$)} \\ 
    \midrule
    \multirow{4}{*}{Chain-of-thought (CoT) \cite{12(wei2022chain)}} &
    \multirow{2}{*}{\textbf{DeepAbduction-Pollution}}
    & 2 & 0.58 & 85.9 & 47.4  \\
    & & 3 & 0.49 & 77.8 & 39.1  \\
    \addlinespace[1pt]
    \cline{2-6}
    \addlinespace[1pt]
    & \multirow{2}{*}{\textbf{DeepAbduction-Disease}} 
    & 2 & 0.51 & 92.3 & 37.1  \\
    & & 3 & 0.47 & 88.5 & 34.3  \\
    \midrule
    \multirow{4}{*}{\textbf{IFAR (Ours)}} &
    \multirow{2}{*}{\textbf{DeepAbduction-Pollution}}
    & 2 & 0.93 & 92.0 & 94.4  \\
    & & 3 & 0.79 & 77.6 & 82.7  \\
    \addlinespace[1pt]
    \cline{2-6}
    \addlinespace[1pt]
    & \multirow{2}{*}{\textbf{DeepAbduction-Disease}} 
    & 2 & 0.94 & 92.9 & 98.1  \\
    & & 3 & 0.88 & 86.7 & 90.3  \\
    \bottomrule
  \end{tabular}
  \label{differ_dep}
\end{table*}

\begin{table*}[htbp]
  \centering
      \caption{ The performance of IFAR and CoT method across different target perspectives, with Deepseek-V3.}
     \begin{tabular}{c|c|c|ccc}
     \toprule
    \textbf{Method} &\textbf{Dataset} & 
    \textbf{Target Perspective} &
    \textbf{F1 Score ($\uparrow$)} & \textbf{Recall (\%) ($\uparrow$)} & \textbf{Precision (\%) ($\uparrow$)} \\ 
    \midrule
    \multirow{7}{*}{Chain-of-thought (CoT) \cite{12(wei2022chain)}} &
    \multirow{4}{*}{\textbf{DeepAbduction-Pollution}} 
    & Technical & 0.55 & 74.6 & 47.9  \\
    & & Social & 0.60 & 85.3 & 48.3  \\
    & & Policy & 0.45 & 71.5 & 35.4  \\
    & & Economic & 0.45 & 90.7 & 32.4  \\
    \addlinespace[1pt]
    \cline{2-6}
    \addlinespace[1pt]
    & \multirow{3}{*}{\textbf{DeepAbduction-Disease}} 
    & Lifestyle & 0.48 & 91.1 & 35.0  \\
    & & Environment & 0.62 & 93.4 & 51.8  \\
    & & Medical Care & 0.59 & 94.2 & 47.0  \\
    \midrule
    \multirow{7}{*}{\textbf{IFAR (Ours)}} &
    \multirow{4}{*}{\textbf{DeepAbduction-Pollution}} 
    & Technical & 0.80 & 76.2 & 88.1  \\
    & & Social & 0.98 & 96.0 & 100.0  \\
    & & Policy & 0.91 & 90.3 & 91.7  \\
    & & Economic & 0.94 & 94.4 & 93.1  \\
    \addlinespace[1pt]
    \cline{2-6}
    \addlinespace[1pt]
    & \multirow{3}{*}{\textbf{DeepAbduction-Disease}} 
    & Lifestyle & 0.90 & 89.0 & 92.0  \\
    & & Environment & 0.98 & 96.7 & 100.0  \\
    & & Medical Care & 0.93 & 97.4 & 92.3  \\
    \bottomrule
  \end{tabular}
  \label{differ_tar}
\end{table*}

\subsection{Discussions}
\paragraph{Ablation Results}

We remove FCV from our method and observe that, with only ICS, recall increases (e.g., 87.3 $\to$ 94.7 on DeepAbduction-Pollution and 88.7 $\to$ 93.5 on DeepAbduction-Disease with Deepseek-V3), precision decreases (e.g., 98.5 $\to$ 88.2 on DeepAbduction-Pollution and 99.3 $\to$ 87.9 on DeepAbduction-Disease with Deepseek-V3), and F1 score also decrease (e.g., 0.93 $\to$ 0.91 on DeepAbduction-Pollution and 0.94 $\to$ 0.91 on DeepAbduction-Disease with Deepseek-V3), as shown in Table \ref{table2}. It is evident that although the FCV validation can detect incorrect content, thus benefiting precision, it may also mistakenly classify correct content as incorrect, which affects recall. It is worth noting that the key rule is that precision and recall are considered equally important, and therefore we use F1 score as the metric. Based on this, FCV shows a beneficial effect, with its positive impact on precision outweighing its negative impact on recall. If precision and recall were assigned different weights, the results might vary.

We also incorporate the FCV module into the IO method and evaluate its effectiveness. As shown in Table \ref{table2}, "IO and FCV" does not yield a significant performance improvement over the IO method. This is because the IO method itself already has a relatively low recall rate, with many correct answers missing before the validation, which diminishes the impact of the FCV module.

\paragraph{Case Study}
We investigate the reasons behind the failure of the IO method through examples, as shown in Figure \ref{fig6}, in which the goal is to find the second-level causes of water pollution from social perspective. There is no error in identifying the first-level causes (shown in green), but a hallucination error occurs for the second-level causes (shown in orange). This shows that while LLMs can correctly find the simpler direct causes, they struggle with cross-level reasoning and provide no mechanism for correction. 

To mitigate these risks, we simplify the reasoning process by focusing on direct causes at each step in the ICS module and introduce the FCV module to further identify and eliminate the errors in ICS and determine whether the current cause aligns with the target depth and perspective. Figure \ref{fig10} presents a detailed example demonstrating how the two models within our proposed framework interact and cooperate progressively to accomplish the task.

\paragraph{Performance on Different Levels and Perspectives}
We conducted separate statistics for different reasoning levels and perspectives, and the results are shown in Table \ref{differ_dep} and Table \ref{differ_tar}. As the reasoning level deepens, the various metrics of IFAR show a decline, which is reasonable since deeper reasoning is more challenging. For different causal perspectives, the results vary, which may be related to the inherent difficulty of each perspective and the varying extent of the LLM's knowledge across different perspectives. However, in all scenarios, the F1 score of our proposed IFAR is significantly higher than that of the CoT. The recall rate of CoT is relatively close to our method, and in some cases even slightly higher, but there is a significant gap in precision, which is consistent with the conclusion in the main results.

\begin{figure}[ht]
\centering
    \includegraphics[width=1.0\linewidth]{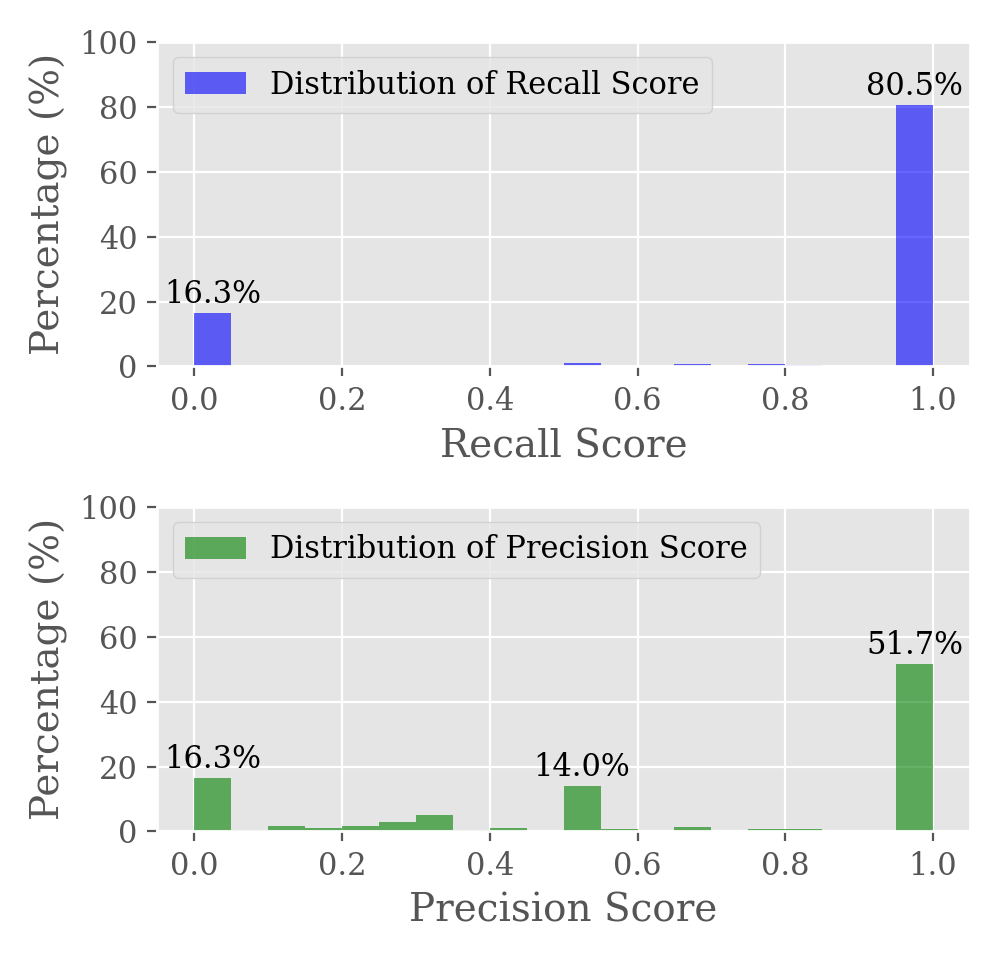}
    \caption{The distribution of the recall and precision for each call of the ICS module in IFAR. The data in the figure represents the statistical results obtained by combining the results from GLM-4-Air, Qwen-2.5-14B, Deepseek-V3 and Deepseek-R1 on the two datasets in DeepAbduction.}
    \label{fig4}
\end{figure}

\paragraph{Analysis of Inverse Cause Seeker (ICS)}
As shown in Figure \ref{fig4}, we further analyze the recall and precision rates of the ICS module during each call on GLM-4-Air, Qwen-2.5-14B, Deepseek-V3 and Deepseek-R1, which includes intermediate reasoning processes, not limited to the final result. It can be observed that the ICS module performs well in terms of recall but exhibits relatively low precision. Relying solely on the ICS module makes it difficult to achieve a balance between recall and precision. The relatively low precision rates in the "Only ICS" row of Table \ref{table2} support this conclusion. Therefore, effective methods to improve precision are necessary, which is why we designed the FCV module to follow the ICS. However, ICS module also plays an important role. With the good recall performance of the ICS module, the detection capability of FCV module can be fully demonstrated, ultimately achieving an effective balance between recall and precision. This also explains why "IO and FCV" method performs not well in Table \ref{table2}.

\paragraph{Analysis of Forward Cause Verifier (FCV)}

\begin{figure}[ht]
  \centering
    \subfloat[IFAR method.]{
    \includegraphics[width=1.6in]
    {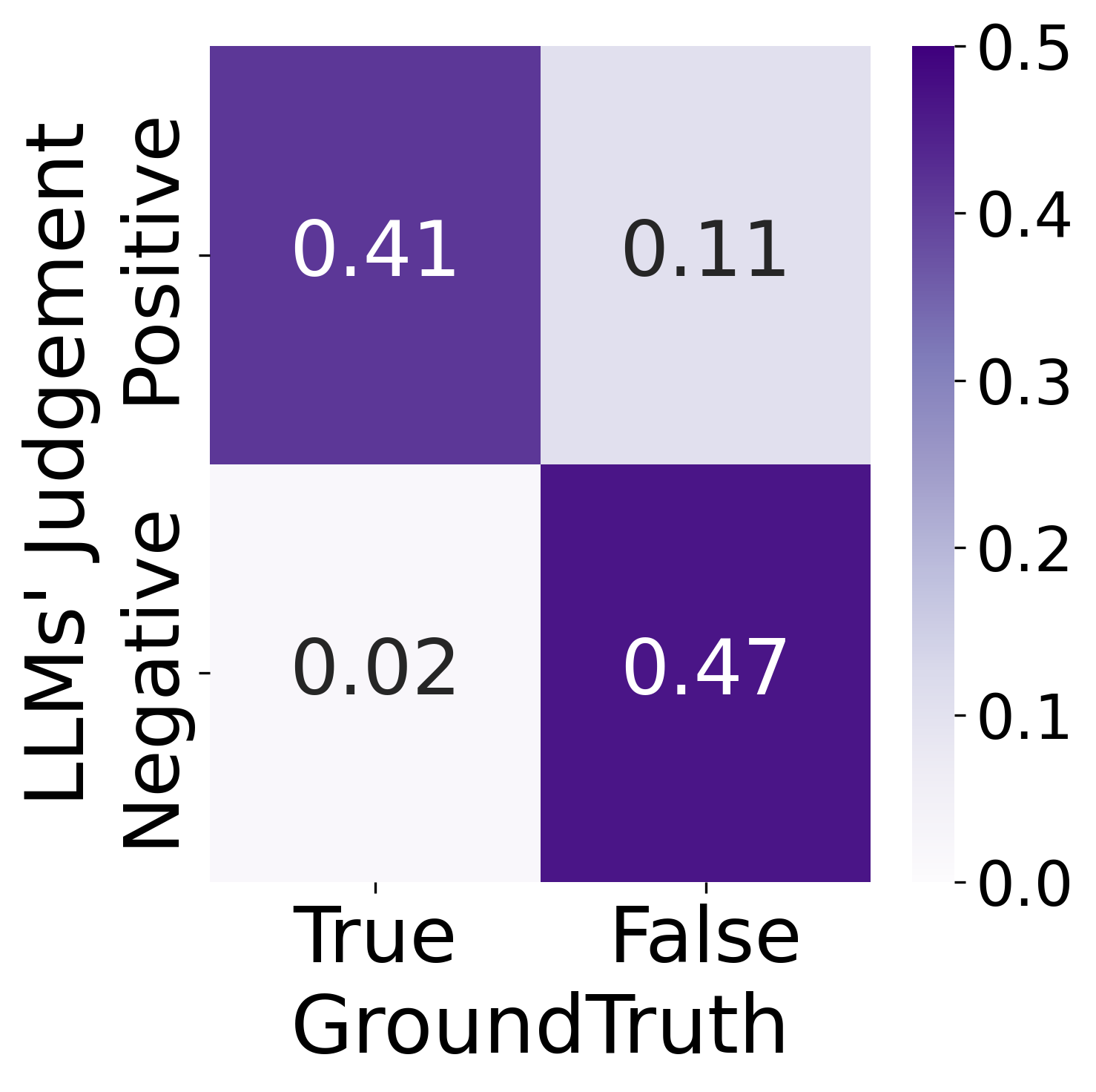}}
    \label{fig:1}
    \subfloat[IO + FCV method.]{
    \includegraphics[width=1.6in]{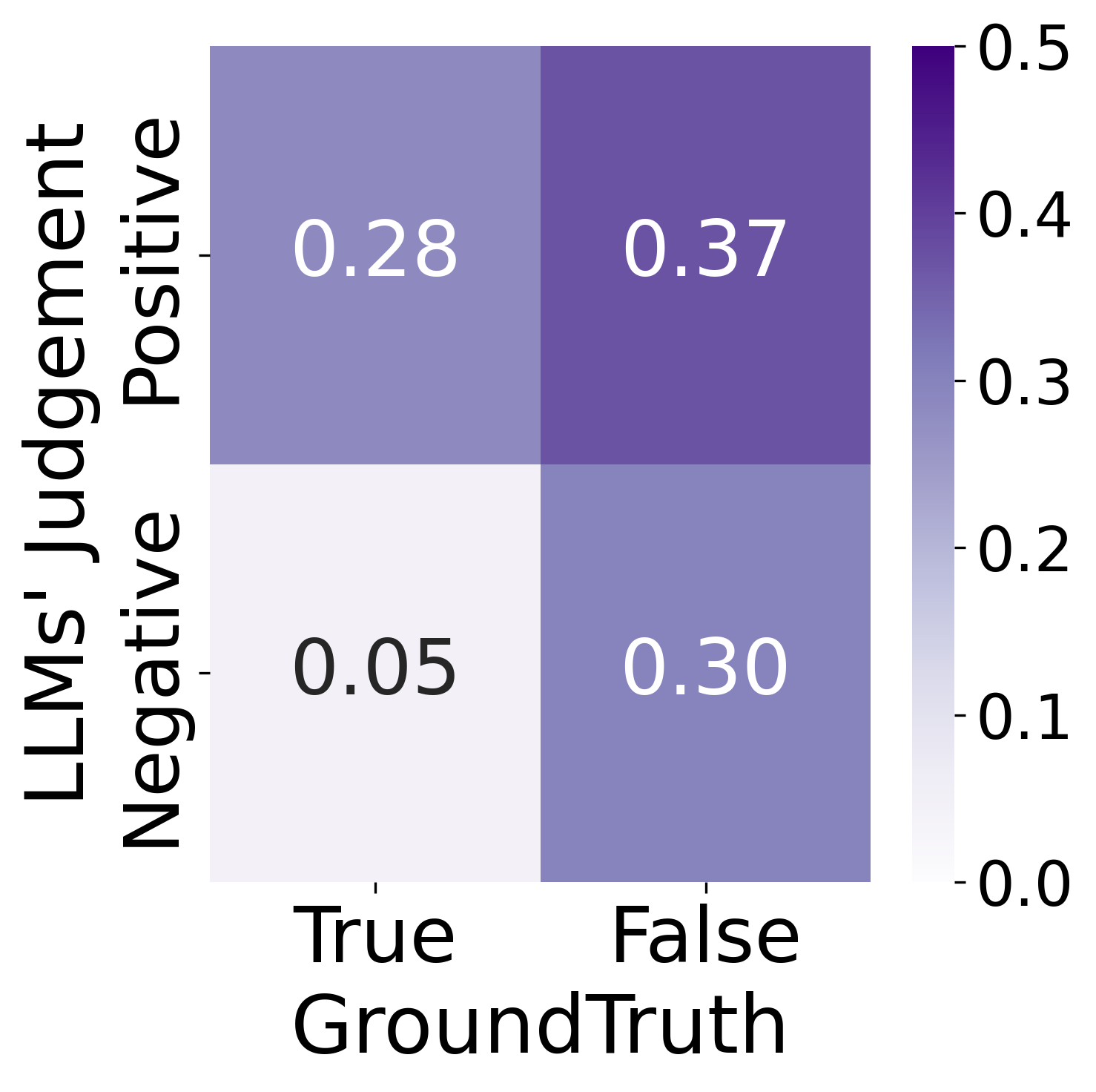}}
    \label{fig:2}
  \caption{We collect the results of each call to the FCV in IFAR and "IO and FCV" method. We use confusion matrixs to represent their performance, where x-axis represents the ground truth and y-axis represents the predicted results. True-Positive indicates the rate that model predicted positive and actual value is also true. Definitions of False-Positive, True-Negative and False-Negative are similarly.}
  \label{fig:two_images}
\end{figure}

We analyze the performance of FCV module under two methods: IFAR and "IO and FCV". As shown in Figure \ref{fig:two_images}, the FCV module performs well in IFAR, with both True-Positive and False-Negative cases having a high proportion. The errors mainly occur when some false instances are incorrectly classified as true (0.11 for the False-Positive). However, the misclassification of correct samples is relatively rare (0.02 for the True-Negative). This is an interesting phenomenon, as it ensures that the impact of FCV module on recall rate is minimal. Thus, adding FCV module to ICS can retain the good results of recall to a large extent, and on this basis, filter out incorrect content to improve precision.

Figure \ref{fig:two_images} shows that adding FCV module to IO method does not perform as well as in the IFAR. With IO method, FCV tends to misclassify the false values as correct (0.37 for the False-Positive). The reason is that in IO and IFAR methods, the reasoning levels differ at each step. IO attempts to directly infer the final answer, so the cause and phenomenon are typically two or three steps apart. In contrast, IFAR searches for direct causes at each step, making it a one-step reasoning. the FCV module is effective in validating one-step reasoning, but it still cannot effectively distinguish errors in multi-step reasoning. The True-Negative still remains low. In summary, the FCV module is more suitable for validating one-step reasoning, which can lead to a greater improvement in precision. It consistently shows that True-Negative errors are minimal, leading to only a slight reduction in recall. Therefore, FCV module has a generally positive impact on the F1 score.

\section{Limitations and Future Directions}
Firstly, DeepAbduction is not rich enough. While the selected themes of pollution and disease are among the most prominent and illustrative scenarios for abductive reasoning, they do not encompass the full spectrum of possibilities. To address this, we plan to enhance our dataset generation methods to cover a wider variety of themes in future work.

Secondly, we treat recall and precision as equally important (as the definition of F1 score), without considering the method adaptation for conditions that recall and precision have different importance. The fine-grained control over the depth and number of causes in a single step of reasoning or validation maybe works. When placing greater emphasis on recall, to prevent correct samples from being misclassified, stricter validation requirements can be imposed. Conversely, when placing more emphasis on precision, the opposite approach can be taken by relaxing the validation criteria.

\section{Conclusion}

In this paper, we explore the abductive reasoning problem in LLMs from multiple perspectives and multiple levels. As the first work to focus on this task, we construct a dataset for multi-perspective and multi-level abductive reasoning, named DeepAbduction. We also propose the \textsc{Inverse-Forward Abductive Reasoning} (IFAR) framework on this task for LLMs. IFAR is a zero-shot method, consisting of two modules, Inverse Cause Seeker (ICS) and Forward Cause Verifier (FCV). Through experiments and discussions, we demonstrate that our framework outperforms existing methods while maintaining a balance between recall and precision metrics. Furthermore, IFAR can enhance the performance of a non-reasoning type LLM to surpass that of reasoning-type LLMs, and it remains effective even when used on the latter. We hope that our work will inspire future studies, particularly in the field of LLMs reasoning, and contribute to the advancement of intelligent reasoning.

\bibliographystyle{IEEEtran}
\bibliography{TTTTTAI}

@ARTICLE{tai1,
  author={Ban, Taiyu and Chen, Lyuzhou and Lyu, Derui and Wang, Xiangyu and Zhu, Qinrui and Tu, Qiang and Chen, Huanhuan},
  journal={IEEE Transactions on Artificial Intelligence}, 
  title={Integrating Large Language Model for Improved Causal Discovery}, 
  year={2025},
  volume={6},
  number={11},
  pages={3030-3042},
  doi={10.1109/TAI.2025.3560927}}

@ARTICLE{tai2,
  author={Mosquera, Manuel and Pinzón, Juan Sebastian and Fonseca, Yesid and Ríos, Manuel and Quijano, Nicanor and Giraldo, Luis Felipe and Manrique, Rubén},
  journal={IEEE Transactions on Artificial Intelligence}, 
  title={Can LLM-Augmented Autonomous Agents Cooperate? An Evaluation of Their Cooperative Capabilities Through Melting Pot}, 
  year={2025},
  volume={},
  number={},
  pages={1-10},
  doi={10.1109/TAI.2025.3569192}}

@ARTICLE{tai3,
  author={Lv, Liuzhenghao and Lin, Zongying and Li, Hao and Liu, Yuyang and Cui, Jiaxi and Chen, Calvin Yu-Chian and Yuan, Li and Tian, Yonghong},
  journal={IEEE Transactions on Artificial Intelligence}, 
  title={ProLLaMA: A Protein Large Language Model for Multi-Task Protein Language Processing}, 
  year={2025},
  volume={},
  number={},
  pages={1-12},
  doi={10.1109/TAI.2025.3564914}}

@ARTICLE{tai4,
  author={Hagos, Desta Haileselassie and Battle, Rick and Rawat, Danda B.},
  journal={IEEE Transactions on Artificial Intelligence}, 
  title={Recent Advances in Generative AI and Large Language Models: Current Status, Challenges, and Perspectives}, 
  year={2024},
  volume={5},
  number={12},
  pages={5873-5893},
  doi={10.1109/TAI.2024.3444742}}

@ARTICLE{tai5,
  author={Liao, Haicheng and Kong, Hanlin and Wang, Bonan and Wang, Chengyue and Ye, Wang and He, Zhengbing and Xu, Chengzhong and Li, Zhenning},
  journal={IEEE Transactions on Artificial Intelligence}, 
  title={CoT-Drive: Efficient Motion Forecasting for Autonomous Driving with LLMs and Chain-of-Thought Prompting}, 
  year={2025},
  volume={},
  number={},
  pages={1-15},
  doi={10.1109/TAI.2025.3564594}}

\end{document}